\documentclass{article}

\usepackage{fullpage}
\usepackage[utf8]{inputenc} 
\usepackage[T1]{fontenc}    
\usepackage{hyperref}       
\usepackage{url}            
\usepackage{booktabs}       
\usepackage{amsfonts}       
\usepackage{nicefrac}       
\usepackage{microtype}      

\usepackage{amssymb,amsmath,amsthm}
\usepackage{color}
\usepackage{verbatim}
\usepackage{bbm}
\usepackage{multirow}
\usepackage{paralist}
\usepackage{nicefrac}
\usepackage{graphicx}

\usepackage{caption}
\usepackage{subcaption}

\newtheorem{theorem}{Theorem}
\newtheorem{prop}[theorem]{Proposition}

\newtheorem{definition}[theorem]{Definition}
\newtheorem{coro}[theorem]{Corollary}

\def\R{\mathbb{R}}

\def\suchthat{\;:\;}
\def\given{\mid}

\newcommand{\abs}[1]{\left|#1\right|}

\newcommand{\norm}[1]{\left\|#1\right\|_{2}}
\newcommand{\ind}[1]{\mathbbm{1}\left(#1\right)}

\newcommand{\prob}[1]{\operatorname{Pr}\left(#1\right)}
\newcommand{\argmin}[1]{\underset{#1}{\operatorname{argmin}}}

\newcommand{\expec}[2]{\mathbbm{E}_{#1}\left[#2\right]}

\title{Rawlsian Fair Adaptation of Deep Learning Classifiers}

\author{%
  Kulin Shah\footnote{Microsoft Research, Email: \texttt{kulin.shah98@gmail.com}}
  \and 
  Pooja Gupta\footnote{Indian Institute of Science, Email: \texttt{poojagupta@iisc.ac.in}}
  \and 
  Amit Deshpande\footnote{Microsoft Research, Email: \texttt{amitdesh@microsoft.com}}
  \and
  Chiranjib Bhattacharyya\footnote{Indian Institute of Science, Email: \texttt{chiru@iisc.ac.in}}   
}
\date{}

\begin{document}
\maketitle

\begin{abstract}
Group-fairness in classification aims for equality of a predictive utility across different sensitive sub-populations, e.g., race or gender. Equality or near-equality constraints in group-fairness often worsen not only the aggregate utility but also the utility for the least advantaged sub-population. In this paper, we apply the principles of Pareto-efficiency and least-difference to the utility being accuracy, as an illustrative example, and arrive at the \emph{Rawls classifier} that minimizes the error rate on the worst-off sensitive sub-population. Our mathematical characterization shows that the \emph{Rawls classifier} uniformly applies a threshold to an ideal \emph{score} of features, in the spirit of fair equality of opportunity. In practice, such a score or a feature representation is often computed by a black-box model that has been useful but unfair. Our second contribution is practical \emph{Rawlsian fair adaptation} of any given black-box deep learning model, without changing the score or feature representation it computes. Given any score function or feature representation and only its second-order statistics on the sensitive sub-populations, we seek a threshold classifier on the given score or a linear threshold classifier on the given feature representation that achieves the \emph{Rawls error rate} restricted to this hypothesis class. Our technical contribution is to formulate the above problems using ambiguous chance constraints, and to provide efficient algorithms for Rawlsian fair adaptation, along with provable upper bounds on the Rawls error rate. Our empirical results show significant improvement over state-of-the-art group-fair algorithms, even without retraining for fairness.
\end{abstract}

\section{Introduction}
\label{sec:introduction}
Algorithmic decisions and risk assessment tools in real-world applications, e.g., recruitment, loan qualification, recidivism, have several examples of effective and scalable black-box models, many of which have received strong criticism for exhibiting racial or gender bias \cite{Angwin2016ProPublica,Barocas2016}. Anti-discrimination laws and official public policy statements \cite{CivilRightsAct,Magarey2004,WhiteHouseReport,ACMStmt} often explicitly demand that such black-box models be fairness-compliant.

Fair classification is an important problem in fairness-aware learning. Previous work on fair classification has largely focused on group-fairness or \emph{classification parity}, which means equal or near-equal predictive performance on different sensitive sub-populations, e.g., similar accuracy \cite{Dieterich2016,Menon2018TheCO,Celis2019ClassificationWF}, statistical parity \cite{Dwork2012FairnessTA}, equalized odds \cite{Hardt2016EqualityOO}, similar false positive rates \cite{Hardt2016EqualityOO}. A popular objective in group-fair classification is accuracy maximization subject to equal or near-equal predictive performance on different sensitive sub-populations.

What makes any fairness objective well founded? In this paper, we consider the basic principles of \emph{Pareto-efficiency} and \emph{least-difference}, following the work of Rawls on distributive justice \cite{Rawls1971, Rawls-JAF-1, Rawls-JAF-2}. A \emph{Pareto-efficient} solution maximizes some aggregate utility over the entire population. Thus, it is not possible to improve the performance on any one sub-population without sacrificing performance on another sub-population beyond what a Pareto-efficient solution gives. The \emph{least-difference} principle allows certain inequalities as fair, if reducing these inequalities either adversely affects the others or does not uplift the worst-off. For example, a group-fair solution that equalizes utilities across different sensitive sub-populations satisfies the least-difference principle vacuously because it has no inequality. Therefore, typical group-fair algorithms maximize accuracy subject to group-fairness constraints ensuring near-equal utilities for different sensitive sub-populations. However, none of the above solutions satisfy both the Pareto-efficiency and least-difference principles simultaneously. Do there exist solutions that satisfy both these principles simultaneously?

The most accurate classifier is the Bayes classifier that predicts using a uniform threshold of $\nicefrac{1}{2}$ applied to the true outcome probability of every example. Previous work provides similar characterization of optimal group-fair classifiers as threshold classifiers on some ideal score \cite{Menon2018TheCO,Celis2019ClassificationWF,CorbettDavies2017AlgorithmicDM,CorbettDavies2018TheMA}. Do such threshold-on-score characterizations hold for Rawlsian fairness too? 

Most black-box models in practice are often optimized for accuracy, and they output a risk score or a feature representation along with their prediction. Even when an existing model is found to be biased during a fairness audit, retraining a different fair model is difficult on proprietary or private training data. Algorithmic ideas to develop group-fair classifiers that maximize accuracy subject to group-fairness constraints require estimates of true outcome probability \cite{Menon2018TheCO,Celis2019ClassificationWF}, which are intractable with finite samples. Most post-processing methods for group-fairness also require estimates of the true outcome probabilities \cite{Pleiss2017OnFA,Hardt2016EqualityOO,Kamiran2012DecisionTF}. Secondly, existing group-fairness toolkit \cite{Celis2019ClassificationWF} can use a given risk score as an input feature and adapt for group-fairness but the resulting group-fair classifier does not predict using a threshold on that same risk score. Is fair adaptation possible without changing the risk scores or feature representations?

The cost or benefit of a decision for an individual depends on the protected attribute (e.g., race, gender, age) as well as the true class label. For example, an underprivileged person who qualifies for a loan and successfully repays, has a greater utility than a privileged person who qualifies for the same loan and repays, and also greater utility than another underprivileged person who qualifies but does not repay. Many group-fair classifiers with high and near-equal accuracy on different races or genders can have poor accuracy on either the positive or negative class therein. This typecasting is exacerbated when there is class imbalance within a group. Therefore, we consider the sensitive sub-populations defined by group memberships (e.g., race, gender, age) as well as true class labels.

\subsection{Our results}
In Section \ref{sec:rawls}, we define the \emph{Rawls classifier} as an optimal solution to an objective that simultaneously satisfies both the \emph{Pareto-efficiency} and \emph{least-difference} principles in fair classification. When the classification utility is quantified by accuracy, the \emph{Rawls classifier} minimizes the error rate on the worst-off sensitive sub-population over all classifiers; we call its optimal value as the \emph{Rawls error rate} (see Definition \ref{def:rawls}). The max-min objective is well-known in social choice theory for equitable distribution of goods \cite{hammond1976equity,strasnick1976social}; we provide a formulation so that it applies to fair classification. We mathematically characterize the \emph{Rawls classifier} and the \emph{Rawls error rate} (Theorem \ref{thm:rawls}). Moreover, our additional observations in Subsection \ref{subsec:rawls-prop} show that the characterization of the \emph{Rawls classifier} reveals interesting, non-trivial properties about its most disadvantaged sensitive sub-populations.

Our mathematical characterization of the \emph{Rawls classifier} shows that it uniformly applies a threshold to a certain ideal \emph{score} of features, whose description requires the underlying data distribution explicitly. We give a description of the \emph{Rawls classifier} as a threshold on an ideal \emph{score} function given by a convex combination of signed \emph{unveil functions}, which quantify the likelihood of an individual belonging to a sensitive sub-population given only the unprotected attributes. Computing these unveil functions from a finite sample drawn randomly from an arbitrary underlying distribution is intractable, a familiar obstacle encountered for exact implementation of the accuracy maximizing Bayes classifier and the optimal group-fair classifier characterized in previous work \cite{Menon2018TheCO,Celis2019ClassificationWF}.

For Rawlsian \emph{fair adaptation} of deep learning classifiers in practice, we consider the following formulation. Given a non-ideal score function or a feature map computed by a black-box model, and only its second-order statistics on the sensitive sub-populations, we seek a threshold classifier on the given non-ideal score or a linear threshold classifier on the given feature map that achieves the Rawls error rate in this restricted hypothesis class. In Section \ref{sec:fair-adapt}, we formulate the above problems using ambiguous chance constraints, and provide efficient algorithms for fair adaptation, along with provable upper bounds on the Rawls error rate in the restricted setting. In Subsection \ref{subsec:flat}, we show that, when the feature map distributions conditioned on each sensitive sub-population are Gaussian and we seek a linear threshold classifier, then we can provably and efficiently achieve the restricted Rawls error rate (Theorem \ref{thm:flat}).

In Section \ref{sec:experiments}, we show that our Rawlsian fair adaptation formulation above is readily applicable to any black-box model that computes a score function or a feature representation. For example, we train a model to maximize classification accuracy on a standard dataset used in text classification for toxicity. Our Rawlsian adaptation using its label-likelihood scores and feature representation does not require retraining a different fair model, and shows a significant improvement in the error rates on the worst-off sensitive sub-populations. We also show a similar improvement over real-world and synthetic data sets, when compared against best known post-processing fairness methods \cite{Kamiran2012DecisionTF} and group-fair classifiers \cite{Celis2019ClassificationWF} as our baselines.


\subsection{Related work}
In the context of fair classification, recent work by Hashimoto et al. \cite{hashimoto18a} studies Rawlsian fairness for empirical risk minimization, and observes that Rawlsian fairness prevents disparity amplification over time, which may be unavoidable if we insist on near-equal group-wise performance as a group-fairness constraint. 
Recent work has also looked at Rawlsian theory to study the veil of ignorance and inequality measurements \cite{heidari2018,gummadi2019,heidari2019,srivastava2019}, contextual bandits \cite{joseph2016fair}, fair meta-learning \cite{zhao2020primal}, envy-free classification \cite{hossain2020}.

Recent work has also looked at Rawlsian theory to study fairness in different settings of machine learning such as the \emph{veil of ignorance}, moral luck-vs-desert, difference principle etc. to study fair equality of opportunity, inequality measurements, bandit problems \cite{heidari2018,gummadi2019,heidari2019,srivastava2019}. Our max-min objective is a \emph{prioritarian} objective based only on the Pareto-efficiency and the least-difference principles from distributive justice.

Previous work has also proposed different approaches for post-processing to achieve group-fairness on black-box models  \cite{Hardt2016EqualityOO,Agarwal2018ARA,dwork18a, Kamiran2012DecisionTF,feldman2015,beutel17}. However, they do not address fair adaptation of black-box deep learning models without changing their scores or feature representation similar to our work.

\section{Notation}
Let $\mathcal{X}$ be the space of input features, $\{0, 1\}$ be the binary class labels, and $[p] = \{1, 2, \dotsc, p\}$ be the set of protected attributes, e.g., race, gender, age. Any input data distribution corresponds to a joint distribution $\mathcal{D}$ on $\mathcal{X} \times \{0, 1\} \times [p]$. Let $(X, Y, Z)$ denote a random element of $\mathcal{X} \times \{0, 1\} \times [p]$ drawn from the joint distribution $\mathcal{D}$. Let $p_{ij}$ denote $\prob{Y=i, Z=j}$. 
\begin{definition} \label{def:sensitive}
For any $i \in \{0, 1\}, j \in [p]$, define the sensitive sub-population $S_{ij} \subseteq \mathcal{X} \times \{0, 1\} \times [p]$ as $S_{ij} = \{(x, i, j) \suchthat x \in \mathcal{X}\}$. 
\end{definition}
The utility of a decision often depends on the protected attribute as well as the true label. This is implicitly considered in metrics such as false positive rates over groups. Our definition of sensitive sub-populations makes this more explicit. 
\begin{definition} \label{def:unveil}
For any $i \in \{0, 1\}, j \in [p]$, define the unveil function $\eta_{ij}: \mathcal{X} \rightarrow \R_{\geq 0}$ of the sensitive sub-population $S_{ij}$ as
\[
\eta_{ij}(x) = \prob{Y=i, Z=j \given X=x}.
\]
We define the normalized unveil function $u_{ij}: \mathcal{X} \rightarrow \R_{\geq 0}$ as
\[
u_{ij}(x) = \eta_{ij}(x)/p_{ij}, \quad \text{where $p_{ij} = \prob{Y=i, Z=j}$}.
\]
\end{definition}
We call $u_{ij}$ as the normalized unveil function because 
\begin{align*}
& \mathbb{E}_X \left[ \frac{\eta_{ij}(X)}{p_{ij}} \right]  = 1, \qquad \text{for all $i \in \{0, 1\}, j \in [p]$}.
\end{align*}

Let $\mathcal{D}_{ij}$ denote the conditional distribution for $X$ given the class label $Y=i$ and the protected attribute $Z=j$, and $X_{ij}$ denote a random element of $\mathcal{X}$ drawn from $\mathcal{D}_{ij}$. 

As is common in fairness literature, we denote a binary classifier by a function $f: \mathcal{X} \rightarrow \{0, 1\}$. Note that this subsumes both group-aware as well as group-blind classifiers depending on whether the protected attributes also appear in $\mathcal{X}$ or not.

\section{The Rawls classifier}
\label{sec:rawls}
A natural way to measure the cost of a binary classifier $f$ on the sensitive sub-population $S_{ij}$ is by its error rate $r_{ij}(f) = \prob{f(X) \neq Y \given Y=i, Z=j}$.

Here are two well-known basic principles of efficiency and fairness from distributive justice and social choice theory, when we consider classification accuracy as utility.
\begin{enumerate}
\item {\bf Pareto-efficiency principle:} A classifier $f$ is Pareto-efficient, if there exists some $\lambda \in \R_{\geq 0}^{2p}$ such that $f$ minimizes ${\sum_{ij} \lambda_{ij} r_{ij}(f)}$ over all $f: \mathcal{X} \rightarrow \{0, 1\}$. As a consequence, it is not possible to improve the performance on any one sub-population without sacrificing performance on another sub-population.
\item {\bf Least-difference principle:} A classifier $f$ satisfies the least-difference principle, if we have $\abs{r_{ij}(g) - r_{kl}(g)} < \abs{r_{ij}(f) - r_{kl}(f)}$, for any classifier $g$ and two sensitive sub-populations $S_{ij}$ and $S_{kl}$, then either $r_{ab}(g) > r_{ab}(f)$, for some $a, b$, or $\max_{(a, b)} r_{ab}(g) \geq \max_{(a, b)} r_{ab}(f)$. In other words, we cannot uplift the worst-off and reduce existing inequality without adversely affecting others.
\end{enumerate}
Let $\hat{f}$ be the Bayes classifier that maximizes accuracy. Then $\hat{f}$ minimizes the total error $\sum_{ij} p_{ij} r_{ij}(f)$ among all $f: \mathcal{X} \rightarrow \{0, 1\}$, and hence it is Pareto-efficient. However, the Bayes classifier $\hat{f}$ can violate the least-difference principle.

A group-fair classifier that equalizes false positive rates $r_{0j}(f)$'s for all $j$, or false negative error rates $r_{1j}(f)$'s for all $j$, vacuously satisfies the least-difference principle, assuming we restrict ourselves to either only $S_{0j}$'s or only $S_{1j}$'s as sensitive sub-populations, respectively. However, a group-fair solution or even a solution that maximizes accuracy subject to near-equality group-fairness constraints is not necessarily Pareto-efficient. Insisting on equality or near-equality often violates the principle of Pareto-efficiency when the only way to achieve equality makes it worse but equal for individual sensitive sub-populations.

Now we state our objective that defines the Rawls classifier, and we will show later in this section that it satisfies both the Pareto-efficiency and least-difference principles. 

\begin{definition} \label{def:rawls}
Given any joint distribution $\mathcal{D}$ on $\mathcal{X} \times \{0, 1\} \times [p]$, we define the Rawls classifier and the Rawls error rate, respectively, as
\begin{align*}
f^{*} & = \argmin{f: \mathcal{X} \rightarrow \{0, 1\}}~ \max_{i \in \{0, 1\}, j \in [p]} r_{ij}(f), \\
r^{*} & = \min_{f: \mathcal{X} \rightarrow \{0, 1\}}~ \max_{i \in \{0, 1\}, j \in [p]} r_{ij}(f),
\end{align*}
where $r_{ij}(f) = \prob{f(X) \neq Y \given Y=i, Z=j}$, which is the error rate of $f$ on the sensitive sub-population $S_{ij}$.
\end{definition}
It is easy to see that the Rawls classifier satisfies the least-difference principle. Any classifier $g$ that reduces existing inequalities from $f^{*}$ with adversely effects, i.e., keeping $r_{ij}(g) \leq r_{ij}(f)$, for all $i, j$, must either leave the maximum error rate untouched. Otherwise, it would contradict $f^{*}$ being the minimizer of $\max_{(i, j)} r_{ij}(f)$.

On the other hand, Pareto-efficiency of the Rawls classifier requires a short proof (see Subsection \ref{subsec:rawls-char}). Interestingly, this also helps in characterizing it as a threshold classifier on an \emph{ideal} score function.

{\bf Remark:} Under reasonable assumptions on utility functions and choice spaces in social choice theory, it is known that any solution that satisfies Pareto-efficiency and least-difference principles simultaneously must actually be a solution that maximizes the minimum utility across participants \cite{hammond1976equity,strasnick1976social}. More specifically, it is a lex-min solution, i.e., if there exist multiple solutions that maximize the minimum utility across participants then a lex-min solution maximizes the second minimum utility among them, and then the third minimum utility and so on. However, in this paper, we focus only on maximizing the minimum utility, or equivalently,  minimizing the maximum error rate.

We show that the error rate $r_{ij}(f)$ of any binary classifier $f$ on its sensitive sub-population $S_{ij}$ can be expressed as a weighted expectation of $f(X)$, weighted by the normalized unveil function $u_{ij}(X)$ of the sensitive sub-population $S_{ij}$ (see Definition \ref{def:unveil}). As a consequence, each $r_{ij}(f)$ is a linear function of $f$, when the data distribution $\mathcal{D}$ is fixed. This is a known observation already used in previous work \cite{Menon2018TheCO, Celis2019ClassificationWF}. 
\begin{prop} \label{prop:error-expec}
For any binary classifier $f: \mathcal{X} \rightarrow \{0, 1\}$, its error rate $r_{ij}(f)$ on a sensitive sub-populations $S_{ij}$ equals 
\[
r_{ij}(f) = \begin{cases} \mathbb{E}_{X} \left[ f(X)~ u_{0j}(X) \right], & \text{for $i=0$}, \\ 1 - \mathbb{E}_X \left[ f(X)~ u_{1j}(X) \right], & \text{for $i=1$}, \end{cases}
\]
where $p_{ij} = \prob{Y=i, Z=j}$ and $u_{ij}(x)$ the normalized unveil function of $S_{ij}$ (see Definition \ref{def:unveil}).
\end{prop}

For any binary classifier $f$, we express its maximum error rate $r_{ij}(f)$ over all sensitive sub-populations $S_{ij}$ as the maximum over all possible convex combinations of $r_{ij}(f)$'s. As a consequence, $\max_{ij} r_{ij}(f)$ is also a linear function of $f$, when the coefficients in the optimal convex combination and the underlying data distribution $\mathcal{D}$ are fixed.
\begin{prop} \label{prop:convex-rawls}
For any binary classifier $f: \mathcal{X} \rightarrow \{0, 1\}$, the maximum error rate over all sensitive sub-populations $S_{ij}$ equals 
\begin{align*}
\max_{i \in \{0, 1\}, j \in [p]} &  r_{ij}(f) = \max_{\substack{\sum_{ij} c_{ij} \leq 1 \\ c_{ij} \geq 0,~ \forall ij}}~  \expec{X}{f(X) \left(\sum_{i \in \{0, 1\}, j \in [p]} (-1)^{i} c_{ij} u_{ij}(X)\right)} + \sum_{j \in [p]} c_{1j}.
\end{align*}
\end{prop}


\subsection{Characterization of the Rawls classifier} \label{subsec:rawls-char}
Now we are ready to characterize the Rawls classifier. We characterize the Rawls classifier as a threshold classifier on an ideal score function that is expressed as a certain convex combination of signed, normalized unveil functions $u_{ij}(x)$'s. Moreover, we show that the non-zero coefficients in this convex combination actually indicate the maximally disadvantaged or vulnerable sensitive sub-populations $S_{ij}$'s.

\begin{theorem} \label{thm:rawls}
Given any data distribution $\mathcal{D}$ on $\mathcal{X} \times \{0, 1\} \times [p]$, there exist non-negative coefficients $c^{*}_{ij}$, for $i \in \{0, 1\}$ and $j \in [p]$, satisfying $\sum_{i \in \{0, 1\}, j \in [p]} c^{*}_{ij} = 1$, such that the Rawls classifier achieving the Rawls error rate is given by 
\[
f^{*}(x) = \mathbb{I} \left( \sum_{j \in [p]} c^{*}_{1j} u_{1j}(x) - \sum_{j \in [p]} c^{*}_{0j} u_{0j}(x) \geq 0 \right).
\]
and the Rawls error rate is equal to $r^{*} = \sum_{i \in \{0, 1\}, j \in [p]} c^{*}_{ij} r_{ij}(f^{*})$.
\end{theorem}
As an immediate corollary, we get Pareto-efficiency of the Rawls classifier because
\[
f^{*} = \argmin{f: \mathcal{X} \rightarrow \{0, 1\}}{\sum_{i \in \{0, 1\}, j \in [p]} c^{*}_{ij} r_{ij}(f)}.
\]
Our proof is inspired by \cite{Menon2018TheCO}, who introduced similar techniques in the context of group-fairness. The main difference from previous work that uses similar techniques \cite{Menon2018TheCO, Celis2019ClassificationWF} is that the coefficients are for each $i \in \{0, 1\}$ and $j \in [p]$, instead of only $j \in [p]$, and moreover, they have a special meaning as we will show that the non-zero coefficients $c^{*}_{ij}$ indicate the most disadvantaged sensitive sub-populations.

In Theorem \ref{thm:rawls}, the indices of non-zero coefficients $c^{*}_{ij}$'s actually correspond to the sensitive sub-populations that attain the Rawls error rate $r^{*}$, and are therefore, the maximally disadvantaged or vulnerable sensitive sub-populations.

\subsection{Properties of the Rawls classifier} \label{subsec:rawls-prop}
An interesting corollary of the above characterization theorem is that for any Rawls classifier, the maximally disadvantaged sensitive sub-population cannot be unique, unless the Rawls classifier is trivial. 
\begin{coro}
\label{coro:rawls-non-unique}
For any Rawls classifier, the sensitive sub-population $S_{ij}$ that attains the Rawls error rate $r_{ij}(f^{*}) = r^{*}$ cannot be unique, unless the Rawls classifier $f^{*}$ is trivial (i.e., all-zeroes or all-ones).
\end{coro}

Another interesting corollary of Theorem \ref{thm:rawls} is that for any Rawls classifier, there exist at least two sensitive sub-populations, one from each class, that both attain the Rawls error rate.
\begin{coro}
\label{coro:rawls-equal-error-rates}
For any Rawls classifier $f^{*}$, there exist two sensitive sub-populations $S_{0j}$ and $S_{1k}$, one from each class, that both of them attain the Rawls error rate of $r_{0j}(f^{*}) = r_{1k}(f^{*}) = r^{*}$, unless the Rawls classifier is trivial (i.e., all-zeroes or all-ones). \\
Remark: For the above sub-populations, $j$ and $k$ need not be equal.
\end{coro}

It is interesting to study the Rawls classifier even in the case when there are no protected attributes, and the sensitive sub-populations are the positive and negative classes, respectively. This is not standard for group-fairness but nevertheless relevant from a broader fairness perspective, i.e., when $p=1$, the Rawls classifier minimizes the maximum of the false positive rate and the false negative rate. As we shall see, in this special case, Theorem \ref{thm:rawls} implies that the Rawls classifier applies a threshold to $\prob{Y=1 \given X=x|}$, similar to but not the same as the well-known Bayes classifier.  
\begin{coro}
\label{coro:rawls-threshold-classifier}
For $p=1$, the Rawls classifier $f^{*}$ is given by $\ind{\eta(x) \geq t}$, where $\eta(x) = \prob{Y=1 \given X=x}$, and the threshold $t$ is equal to
\[
t = \frac{(c_{01}/p_{01})}{(c_{11}/p_{11}) + (c_{01}/p_{01})}.
\]
\end{coro}


\section{The restricted Rawls classifier and fair adaptation} \label{sec:fair-adapt}
Theorem \ref{thm:rawls} shows a characterization of the Rawls classifier as a threshold classifier on an \emph{ideal} score function, which can be defined using the unveil functions $\eta_{ij}(x)$ of the sensitive sub-populations and the optimal coefficients $c^{*}_{ij}$'s. These require the knowledge of the underlying distribution $\mathcal{D}$ explicitly, in addition to the optimal coefficients $c^{*}_{ij}$'s that are not easy to compute. Moreover, in practice, computing the unveil functions $\eta_{ij}(x)$'s from a finite sample drawn from the distribution $\mathcal{D}$ is intractable. This leads to a natural question of whether there exists a more practical definition of the Rawls classifier and the Rawls error rate, and efficient algorithms to achieve these.

In practice, we often have a non-ideal score function or a feature map, and the classifier we compute needs to be efficient, and comes from a restricted hypothesis class, e.g., threshold classifier on a score or linear threshold classifier on a feature map. However, we do not know the underlying data distribution $\mathcal{D}$ explicitly but only have access to certain statistics of the given score function or feature map on each sensitive sub-population. For simplicity, we assume that we are given reliable estimates of the second-order statistics (i.e., means and second moments) of the score function or feature map on all sensitive sub-populations. This is a reasonable assumption because the second-order statistics can be estimated efficiently from a small sample of points from the underlying data distribution $\mathcal{D}$. Assuming no additional knowledge of $\mathcal{D}$ beyond the second-order statistics, leads to the following definitions of the \emph{set of restricted score-distributions} and the \emph{restricted Rawls classifier}.

\begin{definition} \label{def:restricted-sd}
For any set of means $M =\{\mu_{ij}\}_{i \in \{0, 1\}, j \in [p]}$ with $\mu_{ij}$'s in $\R^{d}$ for all $ij$, and any set of covariance matrices $V = \{\Sigma_{ij}\}_{i \in \{0, 1\}, j \in [p]}$ with $\Sigma_{ij} \in \R^{d \times d}$ for all $ij$, define the set of restricted score-distributions $\mathcal{R}_{MV}$ as all score-distribution pairs $(s, \mathcal{D})$ such that $s: \mathcal{X} \rightarrow \R^{d}$ is any score function (for $d=1$) or any feature map (for $d \geq 2$), and $\mathcal{D}$ is any distribution on some $\mathcal{X} \times \{0, 1\} \times [p]$ such that
\begin{align*}
\expec{X_{ij}}{s(X_{ij})} & = \mu_{ij},~ \text{and} \\
\expec{X_{ij}}{\left(s(X_{ij}) - \mu_{ij}\right) \left(s(X_{ij}) - \mu_{ij})^{T}\right)} & = \Sigma_{ij}, \hspace{5mm} \forall i \in \{0, 1\}, j \in [p],
\end{align*}
where $(X, Y, Z)$ is a random sample from distribution $\mathcal{D}$, and $X_{ij}$ is a random sample $X$ conditioned on $Y=i, Z=j$.
\end{definition}

\begin{definition} \label{def:restricted-rawls}
For any set of restricted score-distributions $\mathcal{R}_{MV}$ as in Definition \ref{def:restricted-sd}, and any hypothesis class $\mathcal{F}$ of classifiers $f: \R^{d} \rightarrow \{0, 1\}$, define the restricted Rawls classifier and the restricted Rawls error rate, respectively, as
\begin{align*}
F^{*} & = \argmin{f \in \mathcal{F}}~ \max_{i \in \{0, 1\}, j \in [p]} R_{ij}(f), \\
R^{*} & = \min_{f \in \mathcal{F}}~ \max_{i \in \{0, 1\}, j \in [p]} R_{ij}(f),
\end{align*}
\begin{align*}
\text{where } R_{ij}(f) & = \sup_{(s, \mathcal{D}) \in \mathcal{R}_{MV}} \prob{f(s(X)) \neq Y \given Y=i, Z=j} \\
& = \sup_{(s, \mathcal{D}) \in \mathcal{R}_{MV}} \prob{f(s(X_{ij})) \neq Y}.
\end{align*}
\end{definition}
Restricted Rawls setting described above restricts the classifier $f$ to be from a given hypothesis class but it is also relaxation in the sense that the distribution $\mathcal{D}$ is allowed to vary as long as the second-order statistics of $s(X_{ij})$'s are fixed.

\subsection{Fair Adaptation of Threshold (FAT) in the restricted Rawls setting} \label{subsec:fat}
In Theorem \ref{thm:fat}, we characterize the restricted Rawls classifier and give a constructive, algorithmic proof for finding it by formulating the problem using ambiguous chance constraints. As a result, we are able to take any existing threshold classifier on some score function, collect the second-order statistics of its score function on all sensitive sub-populations, and efficiently adapt it to a restricted Rawls classifier that we call as Fair-Adapted Threshold (FAT) classifier. Note than many existing classifiers optimized for accuracy and group-fairness come with their own score functions, and we can efficiently adapt their thresholds.
\begin{theorem} \label{thm:fat}
For any set of restricted score-distributions $\mathcal{R}_{MV}$ for $d=1$ given by means $\mu_{ij}$'s and variances $\sigma_{ij}^{2}$'s (as in Definition \ref{def:restricted-sd}), and the hypothesis class $\mathcal{F}$ of threshold classifiers $f_{b}(x) = \ind{s(x) \geq b}$, for the underlying score function, the corresponding restricted Rawls classifier is given by $F^{*}(x) = f_{b^{*}}(x) = \ind{s(x) \geq b^{*}}$ where the threshold $b^{*}$ is equal to
\begin{align*}
b^{*} & = \mu_{1j^{*}} - \sigma_{1j^{*}} \sqrt{\frac{\mu_{1j^{*}} - \mu_{0j^{*}}}{\sigma_{1j^{*}} + \sigma_{0j^{*}}}} = \mu_{0j^{*}} + \sigma_{0j^{*}} \sqrt{\frac{\mu_{1j^{*}} - \mu_{0j^{*}}}{\sigma_{1j^{*}} + \sigma_{0j^{*}}}}, \\
& \text{where} \qquad j^{*} = \underset{j \in [p]}{\text{argmin}}~ \frac{\mu_{1j} - \mu_{0j}}{\sigma_{1j} + \sigma_{0j}}.
\end{align*}

\end{theorem}
The algorithmic version of Theorem \ref{thm:fat}, that first estimates $\mu_{ij}$ and $\sigma_{ij}$ using a finite sample from $\mathcal{D}$ and then computes $b^{*}$ is what we call as Fair Adaptation of Threshold (FAT). We call the corresponding classifier $f_{b^{*}}$ as the \emph{Fair-Adapted Threshold (FAT) Classifier}. 


\subsection{Fair Linear-Adaptation of Thresholds (FLAT) in the restricted Rawls setting} \label{subsec:flat}
In this section, we consider the problem of finding fair adaptation in the restricted Rawls setting where we are a given score or feature map $s: \mathcal{X} \rightarrow \R^{d}$ as a black box, and we have its second order statistics of $s(X_{ij})$, where $X_{ij}$ is a random sample from the sensitive sub-populations $S_{ij}$. We seek a linear threshold classifier on the feature map $s(x)$, so as to achieve the restricted Rawls error rate. Another simplifying assumption we make is to let the distributions $s(X_{ij})$'s be Gaussians with the given means $\mu_{ij}$'s and covariance matrices $\Sigma_{ij}$'s. To compare with the remark made after Definition \ref{def:restricted-rawls}, the Gaussian assumption means that the distributions $s(X_{ij})$'s are completely characterized once their means and covariance matrices are known.

In absence of a given feature map, we can also use the basic features, if $\mathcal{X} \subseteq \R^{d}$. Also note that the even group-fair algorithms that implement Gaussian Naive Bayes (e.g., Meta Fair by Celis et al. \cite{Celis2019ClassificationWF}) also estimate second-order statistics of the data. The second-order statistics can be efficiently estimated from a small sample, in contrast with the unveil functions $\eta_{ij}(X)$ that are known to be intractable from a finite sample.

\begin{theorem} \label{thm:flat}
Let $\mathcal{R}_{MV}$ be any restricted set of score-distributions $\mathcal{R}_{MV}$ given by the means $\mu_{ij} \in \R^{d}$ and covariance matrices $\Sigma_{ij} \in \R^{d \times d}$ of $s(X_{ij})$, for a random sample $X_{ij}$ from the sensitive sub-populations $S_{ij}$, with an underlying score map $s: \mathcal{X} \rightarrow \R^{d}$ and data distribution $\mathcal{D}$. Let the restricted hypothesis class $\mathcal{F}$ be linear threshold classifiers $f_{w, b}(x) = \ind{w^{T} s(x) \geq b}$. Assuming that the distributions of $s(X_{ij})$ are Gaussians, the restricted Rawls classifier $f_{w, b}$ is given by solving the following optimization problem.
\begin{align*}
w^{*} & = \argmin{w} \max_{j} \norm{\Sigma_{1j}^{1/2} w} + \norm{\Sigma_{0j}^{1/2} w} \quad \text{subject to} \\
& w^{T} (\mu_{1j} - \mu_{0j}) = 1,~ \forall j \in [p].
\end{align*}
where the restricted Rawls error rate is given by
\[
r^{*} = 1 - \Phi\left(\min_{j} \frac{(w^{*})^{T}(\mu_{1j} - \mu_{0j})}{\norm{\Sigma_{1j}^{1/2} w^{*}} + \norm{\Sigma_{0j}^{1/2} w^{*}}}\right),
\]
where $\Phi(\cdot)$ is the CDF of the standard normal variable $N(0, 1)$. The optimal threshold for the Rawls classifier is given by
\begin{align*}
b_{*} & = w_{*}^{T} \mu_{0j^{*}} + \Phi^{-1}(1 - r^{*})~ \norm{\Sigma_{0j^{*}}^{1/2} w_{*}} = (w_{*})^{T} \mu_{1j^{*}} - \Phi^{-1}(1 - r^{*})~ \norm{\Sigma_{1j^{*}}^{1/2} w_{*}}, \\
& \text{where} \quad j^{*} = \argmin{j} \frac{(w^{*})^{T}(\mu_{1j} - \mu_{0j})}{\norm{\Sigma_{1j}^{1/2} w^{*}} + \norm{\Sigma_{0j}^{1/2} w^{*}}}
\end{align*}
\end{theorem}
We call the restricted Rawls classifier obtained by algorithmic implementation of Theorem \ref{thm:flat} as Fair Linear Adaptation of Thresholds (FLAT) classifier. In the experiments section, we consider two classifiers \begin{inparaenum} \item FLAT-1, where we approximate $s(X_{ij})$'s by spherical Gaussians, and \item FLAT-2, where we approximate $s(X_{ij})$'s by non-spherical Gaussians.\end{inparaenum}


\section{Experiments} \label{sec:experiments}
\subsection{Experimental setup}
Our baselines for comparison include the given black-box model (neural network trained for maximizing accuracy), meta-fair classifier \cite{Celis2019ClassificationWF}, Reject Option Classifier (ROC) \cite{Kamiran2012DecisionTF}. We use maximum error rate across all sensitive sub-populations as the primary metric for evaluation. We use the acronym FAT for Fair Adaptation of Threshold. We use acronyms FLAT1 and FLAT2 to represent Fair Linear Adaptation of Threshold using spherical covariance matrix and complete covariance matrix, respectively. 

In our experiments, we randomly split every dataset (except wikipedia talk page dataset) into training set (80 \%) and testing set (20 \%).  We perform 10 repititions and report average statistics of all algorithms. We used a multilayer perceptron with 2 hidden layers to get the score for FAT algorithm or feature embedding for FLAT1 and FLAT2. We use 100 neurons in the first hidden layer and choose it from the range of 20 to 100 for the second hidden layer using accuracy. We used adam optimizer with batch size 128. We started the training with 0.005 learning rate with the step decay learning rate scheduler and optimized the parameters of the scheduler to maximize the accuracy. 
 
\subsection{Fair adaptation on real world datasets}
In this section, we illustrate our Rawlsian fair adaptation of any given unfair classifier. We use three real world datasets to show experimental results of our proposed algorithms. The details of these datasets are given below. 

\begin{figure}
    \centering
    \begin{subfigure}[b]{0.35\textwidth}
        \centering
        \includegraphics[scale=0.32]{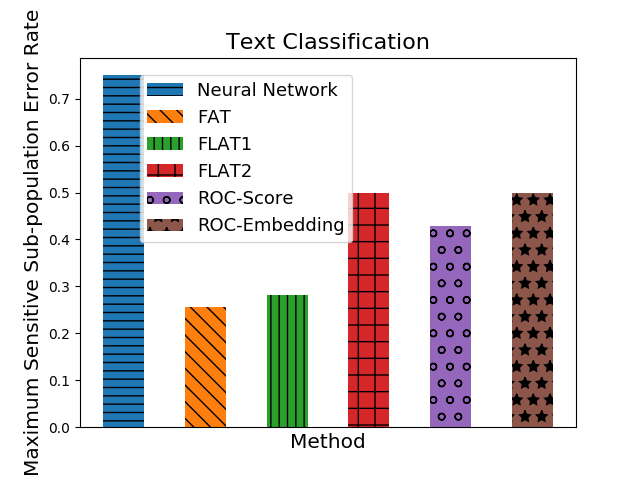}
        \caption{Comparison of maximum sub-population error rate on text classification}
        \label{fig:mgce-text}
    \end{subfigure}%
    \hfill
    \begin{subfigure}[b]{0.3\textwidth}
        \centering
        \includegraphics[scale=0.32]{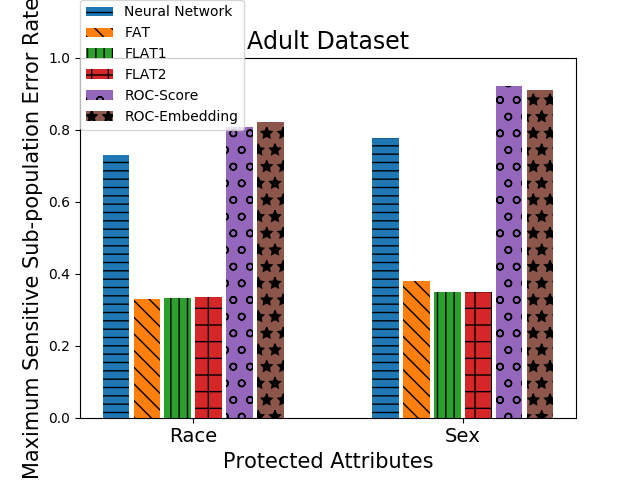}
        \caption{Adaptation of scores or feature embeddings on adult dataset}
        \label{fig:adult-nonlinear-embeddings-comparison}
    \end{subfigure}%
    \hfill
    \begin{subfigure}[b]{0.3\textwidth}
        \centering
        \includegraphics[scale=0.32]{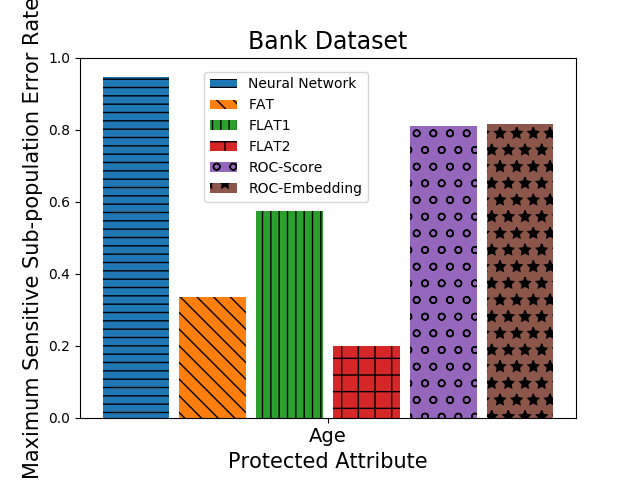}
        \caption{Adaptation of scores or feature embeddings on bank dataset}
        \label{fig:bank-nonlinear-embeddings-comparison}
    \end{subfigure}{}
\end{figure}

\begin{itemize}
\item Wikipedia Talk Page Dataset \cite{wulczyn2017ex}:  In this dataset, the task is to predict whether a comment is toxic. Previous work has pointed out that training to maximize accuracy leads to unintended bias for comments containing terms `gay', `black' etc. (\cite{dixon2018measuring}, \cite{borkan2019nuanced}). The dataset has total 95,691 training examples and 31,867 test examples. We divide all the comments into six groups, that define our protected attributes. First five groups represent whether a comment contains the terms `gay', `islam', `muslim', `male', `black', respectively, and the sixth group contains comments that contain none of the above. Note that a comment can be part of more than one subgroup. 
\item Adult Income Dataset \cite{UCIrepo}: In this dataset, the task is to predict whether an individual has income more than \$50K. The dataset has total 48,842 examples. Protected attributes in the dataset are gender and race. 
\item Bank Dataset \cite{UCIrepo}: In this dataset, the task is to predict whether a client will subscribe to a term deposit. The dataset has total  45,211 examples. Protected attribute in the dataset is age.
\end{itemize}

We first show fair adaptation of non-linear feature embeddings and scores on the above mentioned datasets in Figure \ref{fig:mgce-text}, Figure \ref{fig:adult-nonlinear-embeddings-comparison} and Figure \ref{fig:bank-nonlinear-embeddings-comparison}. On each dataset, we first train a feedforward neural network to maximize overall accuracy. Then, we use second order statistics of scores (for FAT) or feature embeddings (for FLAT1, FLAT2) on each sensitive sub-population for Rawlsian fair adaptation. We use Reject Option Classifier (ROC) applied to the scores (ROC-Score) and feature embeddings (ROC-Embedding) of the neural network for comparison. Please note that ROC-Score and ROC-Embedding use the scores and feature embeddings for the entire data whereas FAT, FLAT1 and FLAT2 only use their second-order statistics on sensitive sub-populations.

In Figure \ref{fig:mgce-text}, we see that FAT and FLAT1 give 10-15 \% improvement in the maximum error rate over all sensitive sub-populations when compared to the baselines. However, FLAT2 has slightly higher maximum sensitive sub-population error rate than ROC-Score because our fair adaptation only uses second-order statistics on sensitive sub-populations. In Figure \ref{fig:adult-nonlinear-embeddings-comparison}, we see that FAT, FLAT1 and FLAT2 achieve around 35-40 \% improvement in the maximum error rate over all sensitive sub-populations when compared to ROC and the given neural network. On the bank dataset (figure \ref{fig:bank-nonlinear-embeddings-comparison}), FAT and FLAT2 achieve around 40-50 \% improvement in the maximum error rate over all sensitive sub-populations when compared to the baselines, whereas FLAT1 achieves around 20 \% improvement over ROC and the given neural network. 

\subsection{Fair adaptation on synthetic datasets}
We show the effectiveness of fair adaptation by showing quantitative and qualitative improvement on synthetic datasets. The synthetic datasets we consider have a binary-valued protected attribute (or two groups) and two classes. We use two-dimensional synthetic datasets for visualization of the decision boundaries given by different classifiers. For each fixed class label and protected attribute value, the features or examples are generated using a spherical Gaussian distribution. The parameters of these Gaussians and number of data points generated are given in Table \ref{tab:synthetic-dataset-1} and Table \ref{tab:synthetic-dataset-2}.
\begin{table}
    \centering
    \begin{tabular}{|c|c|c|c|}
    \hline
        Sub-population & Mean & Variance & Number of Points \\
        \hline
        (0, 0) & (0, -2.5) & 2 & 1900 \\
        \hline
        (0, 1) & (5, 3) & 1 & 100 \\
        \hline
        (1, 0) & (0, 3) & 2 & 1900 \\
        \hline
        (1, 1) & (2, 5) & 1 & 100 \\
    \hline
    \end{tabular}
    \caption{Details of synthetic dataset 1. Sub-population $(i, j)$ means population with label $i$ and protected attribute $j$. }
    \label{tab:synthetic-dataset-1}
\end{table}
\begin{table}
    \centering
    \begin{tabular}{|c|c|c|c|c|}
    \hline
        Sub-population & Mean & Variance & Number of Points \\
        \hline
        (0, 0) & (-5, 0) & 2 & 1900 \\
        \hline
        (0, 1) & (-1, -1) & 1 & 100 \\
        \hline
        (1, 0) & (5, 0) & 2 & 1900 \\
        \hline
        (1, 1) & (1, 1) & 1 & 100 \\
    \hline
    \end{tabular}
    \caption{Details of synthetic dataset 2. Sub-population $(i, j)$ means population with label $i$ and protected attribute $j$. }
    \label{tab:synthetic-dataset-2}
\end{table}

\begin{figure}[t]
    \centering
    \includegraphics[scale=0.32]{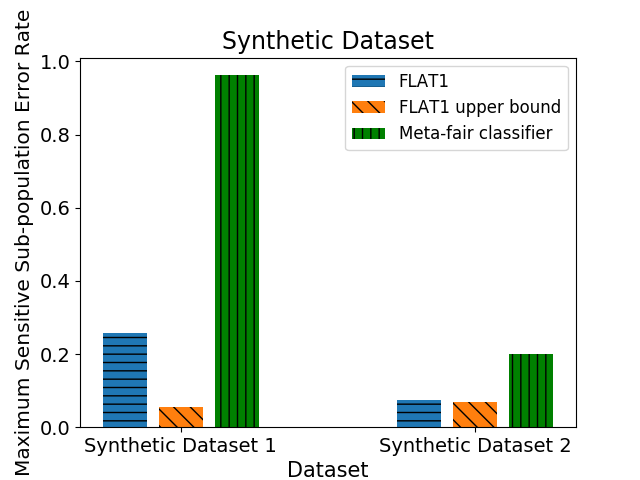}
    \caption{Comparison of maximum sensitive sub-population error rate for synthetic datasets}
    \label{fig:synthetic-bound-comparison}
\end{figure}

We compare the maximum error rate over all sensitive sub-populations for FLAT1 and Meta-fair classifier, and also the upper bound guarantee on this metric for FLAT1 computed by our algorithm in Figure \ref{fig:synthetic-bound-comparison}. We see that FLAT1 outperforms state-of-the-art Meta-fair classifier and achieves around 70 \% smaller value of the maximum error rate over sensitive sub-populations in synthetic dataset 1, and around 15 \% smaller value in synthetic dataset 2.

\begin{figure}
    \centering
    \begin{subfigure}[b]{0.45\textwidth}
        \centering
        \includegraphics[scale=0.5]{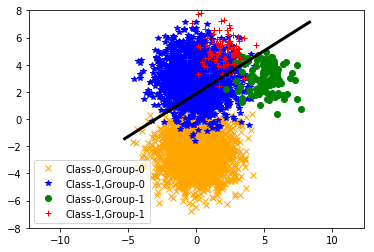}
        \caption{FLAT1 algorithm}
        \label{fig:decision-boundary-synthetic1-FLAT1}
    \end{subfigure}%
    \hfill
    \begin{subfigure}[b]{0.45\textwidth}
        \centering
        \includegraphics[scale=0.4]{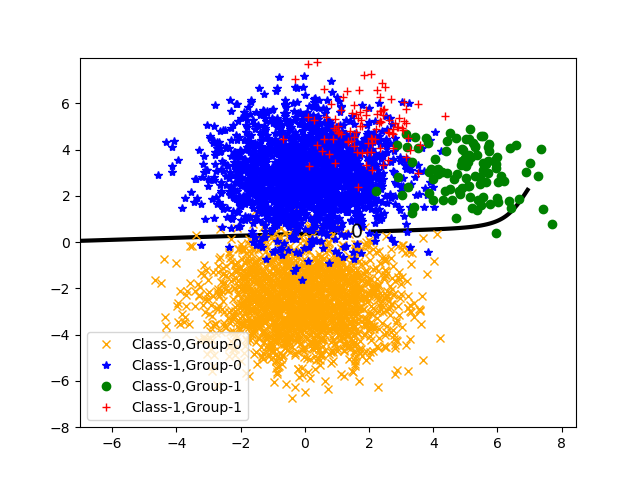}
        \caption{Meta-fair classifier}
        \label{fig:decision-boundary-synthetic1-meta-fair}
    \end{subfigure}
    \caption{Decision boundary (black line) of FLAT1 and Meta-fair classifier on synthetic data 1}
    \label{fig:results-synthetic1}
\end{figure}

\begin{figure}
\centering
    \begin{subfigure}[b]{0.45\textwidth}
        \centering
        \includegraphics[scale=0.5]{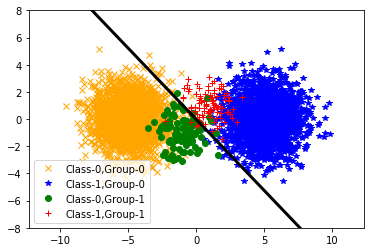}
        \caption{FLAT1 algorithm}
        \label{fig:decision-boundary-synthetic2-FLAT1}
    \end{subfigure}%
    \hfill
    \begin{subfigure}[b]{0.45\textwidth}
        \centering
        \includegraphics[scale=0.42]{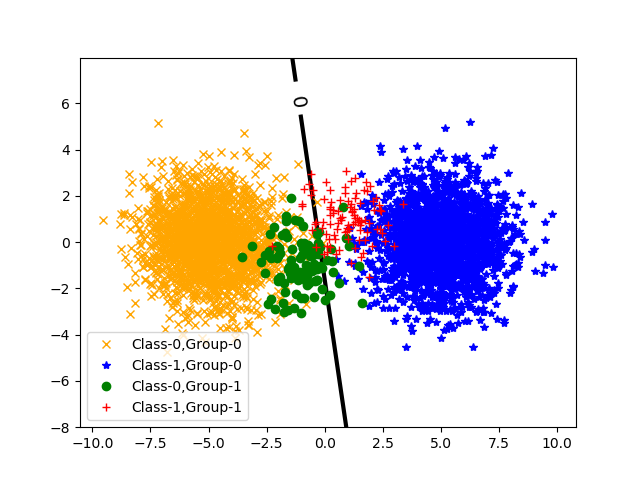}
        \caption{Meta-fair classifier}
        \label{fig:decision-boundary-synthetic2-meta-fair}
    \end{subfigure}
     \caption{Decision boundary (black line) of FLAT1 and Meta-fair classifier on synthetic data 2}
     \label{fig:results-synthetic2}
\end{figure}

Figure \ref{fig:results-synthetic1} and Figure \ref{fig:results-synthetic2} show the different decision boundaries learnt by our fair adaptation and Meta-fair. As Meta-fair classifier defines sub-population using only protected attribute and learns its decision boundary to equalize error rate on each sub-population defined only by the protected attribute, it almost ignores sub-population with label 0 and protected attribute value 1 in synthetic dataset 1. Moreover, Meta-fair classifier learns a non-linear complex classifier and still performs poorly. Because of Rawlsian fairness in our objective and defining the sub-population using class label and protected attribute, FLAT1 learns simple linear classifier and achieves lower maximum error rate across sensitive sub-populations. Even in synthetic dataset 2, FLAT1 learns a tilted decision boundary but Meta-fair learns a vertical decision boundary by ignoring sub-population with protected attribute value 1. Hence, we see that Rawlsian fairness objective and the definition of a sub-population using class label and protected attribute value helps us to do better than group-fair classifiers by only using second order statistics, and address typecasting of sub-populations caused by class imbalance. 

\subsection{Additional fair adaptation results}
Here, we show additional fair adaptation results on two datasets. The details of these datasets are given below.
\begin{itemize}
    \item COMPAS dataset \cite{Compas}: In this dataset, the task is to predict recidivism from an individual's previous history (e.g. previous criminal history, prison time, etc.). The dataset has total 5,278 examples. Protected attributes in the dataset is gender and race. 
    \item German dataset \cite{UCIrepo}: In this dataset, the task is to predict whether an individual is a good credit risk. The dataset has total 1,000 examples. Protected attributes in the dataset is gender and age.
\end{itemize}{}

\begin{figure}
    \centering
    \begin{subfigure}[b]{0.45\textwidth}
        \centering
        \includegraphics[scale=0.37]{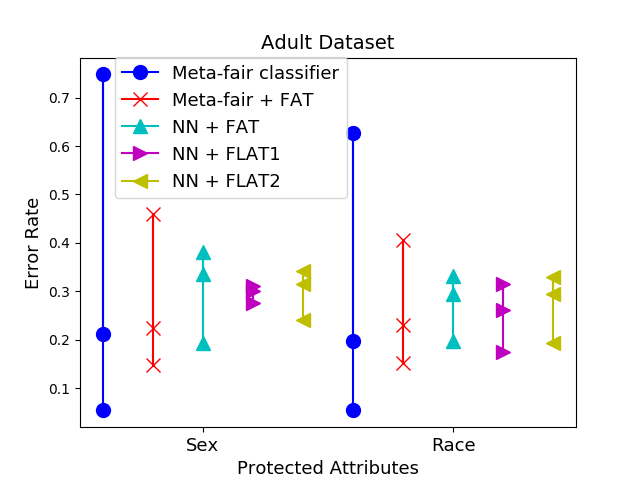}
        \caption{Results on Adult dataset}
        \label{fig:adult-fpr-fnr-acc}
    \end{subfigure}%
    \hfill
    \begin{subfigure}[b]{0.45\textwidth}
        \centering
        \includegraphics[scale=0.37]{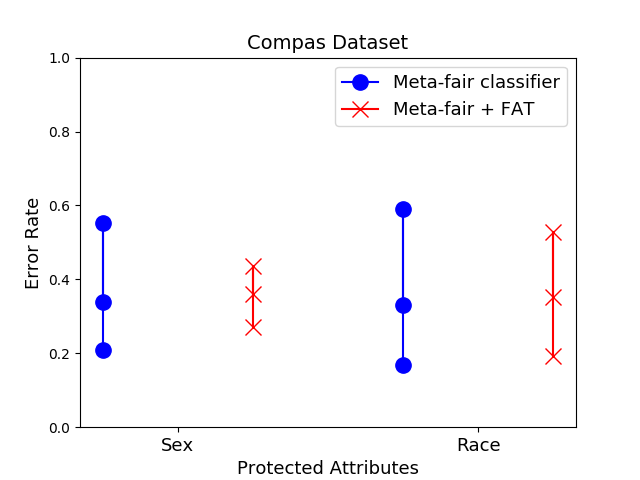}
        \caption{Results on Compas dataset}
        \label{fig:compas-fpr-fnr-acc}
    \end{subfigure}{}
    \caption{Comparison of range of FPR and FNR on Adult and Compas dataset. The lowest and highest point in a line denotes minimum and maximum value of FPR and FNR among all groups, respectively. The middle point among three points in a line denotes error rate.}
    \label{fig:adult-acc}
\end{figure}

In Figure \ref{fig:adult-acc}, we compare range of FPR and FNR of each group on Adult and Compas dataset. We use meta-fair classifier optimized to get near-equal FNR on each group as our baseline. As meta-fair classifier maximizes accuracy on combined data, it ignores one or more sensitive sub-population and ends up getting wide range of FPR and FNR values on each sensitive sub-population.  However, in the proposed approaches, even though we are not directly optimizing FPR and FNR on each sensitive sub-population, we get smaller range compared to meta-fair classifier.

\begin{figure}
    \begin{center}
    \includegraphics[scale=0.29]{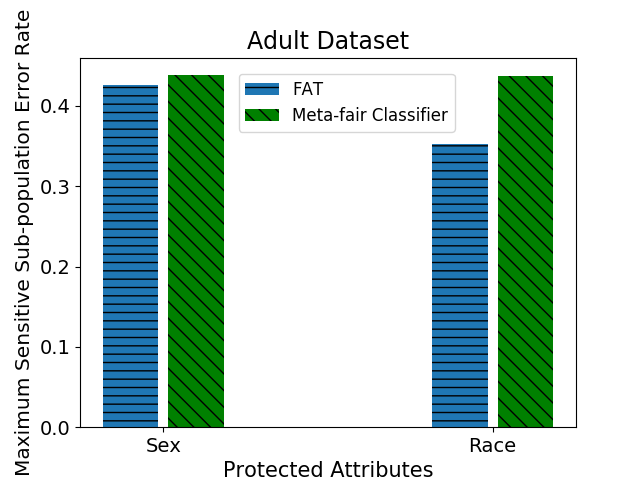} 
    \includegraphics[scale=0.29]{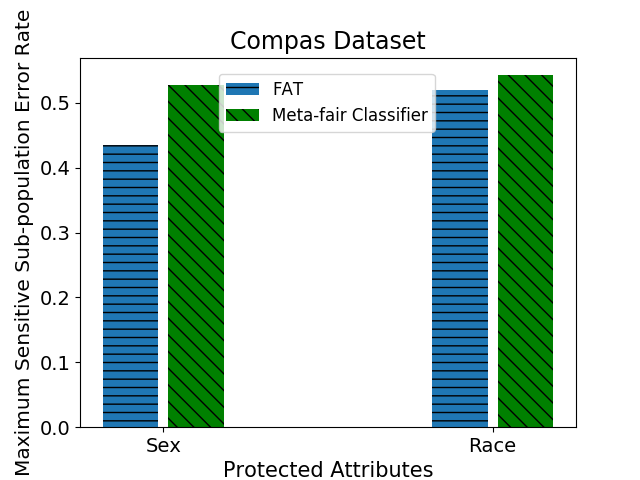} 
    \includegraphics[scale=0.29]{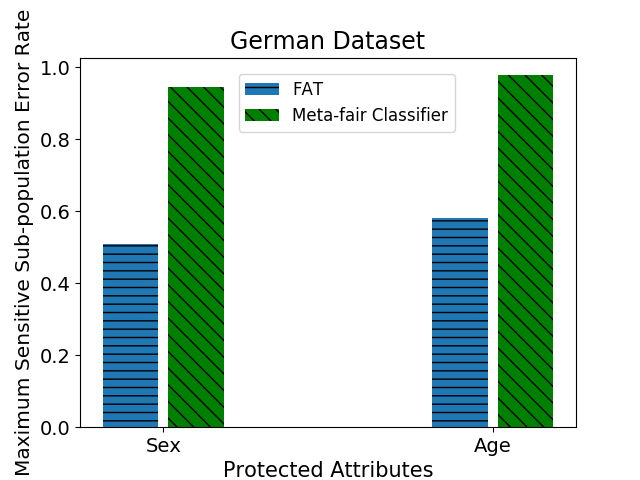}
    \caption{Comparison of FAT algorithm with meta-fair classifier on different datasets and protected attributes} 
    \label{fig:adaptation-fig}
    \end{center}
\end{figure}

We show the results of Fair Adaptation of Threshold (FAT) algorithm in Figure \ref{fig:adaptation-fig} . We use second order statistics of each sub-populations of scores of meta-fair classifier and optimize over threshold to get group fair classifier using FAT algorithm. In Figure \ref{fig:adaptation-fig}, we see that FAT algorithm improves maximum sensitive sub-population error rate by around 5-10 \% in adult dataset and COMPAS dataset but in german dataset, FAT improves maximum sensitive sub-population error rate by around 40-50 \% using only second-order statistics of each sub-populations. 

\begin{figure}
    \centering
    \includegraphics[scale=0.3]{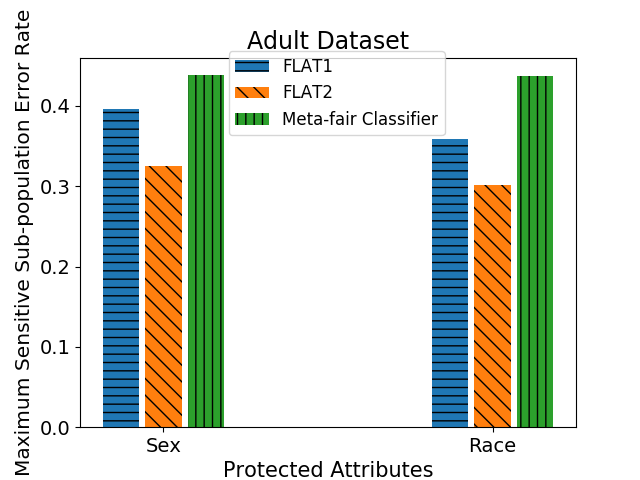} 
    \includegraphics[scale=0.3]{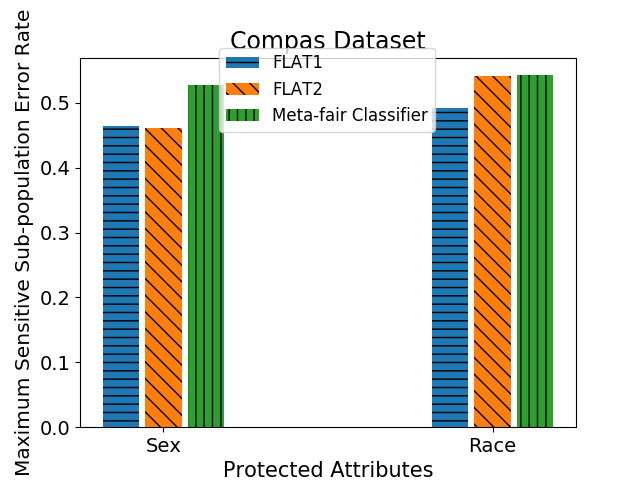}
    \includegraphics[scale=0.3]{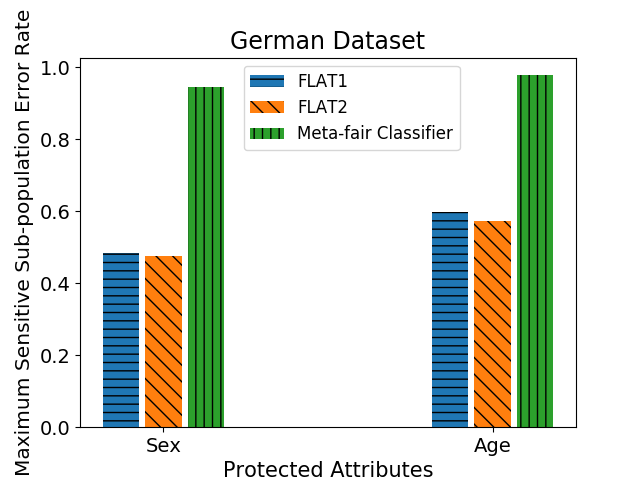}
    \caption{Comparison of FLAT1, FLAT2 algorithm with meta-fair classifier on different datasets and protected attributes}
    \label{fig:real-world-datasets-fig}
\end{figure}

We show experimental results of FLAT1 and FLAT2 on Adult dataset, COMPAS dataset and German Dataset in Figure  \ref{fig:real-world-datasets-fig}. We use second order statistics of features of each sub-population for all the datasets and learn linear classifier on the features using FLAT1 and FLAT2. For comparison, we learn meta-fair classifier using the entire data. 

 From figure \ref{fig:real-world-datasets-fig}, we see that FLAT1 and FLAT2 gets around 10-15 \% lower maximum sensitive sub-population error rate on Adult and COMPAS dataset. In German dataset, FLAT1 and FLAT2 gets around 35-40 \% lower maximum sensitive sub-population error rate. Thus, FLAT1 and FLAT2 achieves lower maximum sensitive sub-population error rate on all the datasets and all the protected attributes using only second order statistics of features of each sub-population. 

\begin{figure}
    \centering
    \includegraphics[scale=0.7]{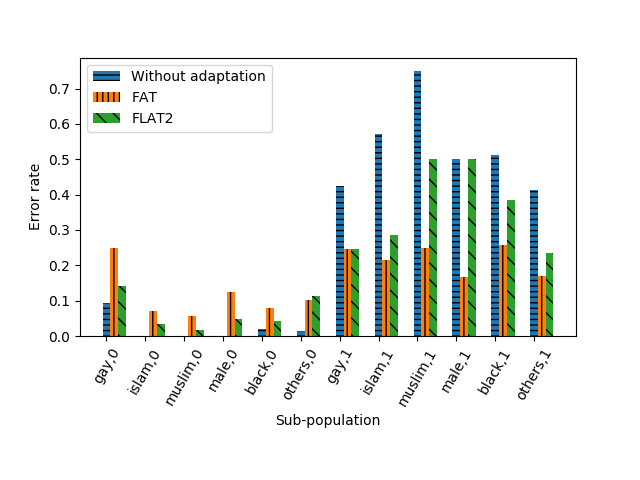}
    \caption{Group-wise class-wise error Rate on all sensitive sub-populations}
    \label{fig:text-groupwise-classwise-error-rate}
\end{figure}

Rawlsian fairness not only tries to minimize maximum sensitive sub-population error rate but in the process, it also decrease gap between error rate among all sensitive sub-populations. In figure \ref{fig:text-groupwise-classwise-error-rate}, we show the comparison of error rate on each sub-population for neural network, FAT and FLAT2. For the figure, it is clear that the proposed algorithm achieves decreases the gap between error rates of sensitive sub-populations. 

{}

\begin{figure}
    \centering
    \includegraphics[scale=0.4]{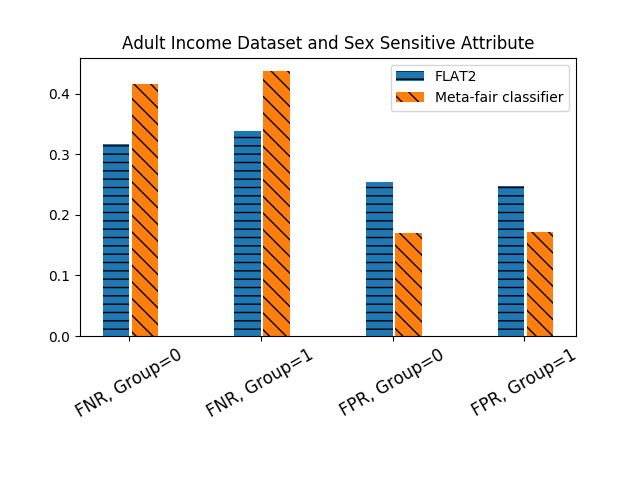}  
    \includegraphics[scale=0.4]{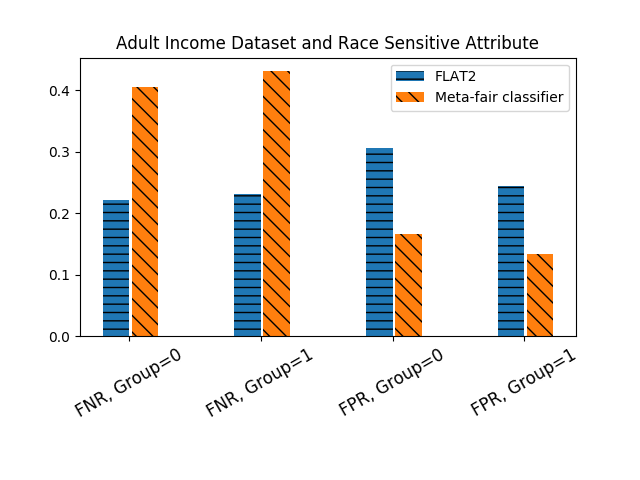} 
    \caption{Comparison of FPR and FNR of FLAT2 algorithm with Meta-fair classifier on Adult dataset}
    \label{fig:adult-fpr-fnr-comparison}
\end{figure}

In figure \ref{fig:adult-fpr-fnr-comparison}, we show experimental results using False Positive Rate (FPR) and False Negative Rate (FNR) metric on Adult dataset. We compare our algorithm with Meta-fair classifier. We see that FLAT2 decreases gap in both metrics (FPR and FNR) between sensitive sub-populations. 

\section{Conclusion}
Starting from the basic principles of Pareto-efficiency and least-difference in distributive justice, we characterize the Rawls classifier for minimizing the maximum error rate across all sensitive sub-populations. We also propose a principled approach for Rawlsian fair adaptation of black-box deep learning models that does not require retraining for fairness, while achieving significant improvement over state of the art group-fair baselines.

\bibliography{main}
\bibliographystyle{plain}

\appendix

\section{Proof details for Section \ref{sec:rawls}}
\label{sec:proof-section-3}
\subsection{Proof for Proposition \ref{prop:error-expec}}
\begin{proof}
The error rate of $f$ on the sensitive sub-population $S_{ij}$ is given by $r_{ij}(f) = \prob{f(X) \neq Y \given Y=i, Z=j}$.
\begin{align*}
\prob{f(X)=1  \given Y=i, Z=j} & = \int_{x \in \mathcal{X}} \ind{f(x) = 1}~ \prob{X=x \given Y=i, Z=j} dx \\
& = \int_{x \in \mathcal{X}} f(x)~ \prob{X=x \given Y=i, Z=j} dx,\\ 
& \hspace{4cm} \text{as $f: \mathcal{X} \rightarrow \{0, 1\}$}  \\
& = \int_{x \in \mathcal{X}} \frac{f(x)~ \prob{Y=i, Z=j \given X=x}~ \prob{X=x}}{\prob{Y=i, Z=j}} dx \\
& = \expec{X}{f(X)~ \frac{\eta_{ij}(X)}{p_{ij}}} \\ &= \expec{X}{f(X)~ u_{ij}(X)}.
\end{align*}
where $\eta_{ij}(x) = \prob{Y=i, Z=j \given X=x}$ and $\eta_{ij}(x)$ and $u_{ij}(x)$ are the unveil function and the normalized unveil function, respectively, of $S_{ij}$ as in Definition \ref{def:unveil}.
\begin{align*}
r_{0j}(f) & = \prob{f(X)=1 \given Y=0, Z=j} \\  
&= \expec{X}{f(X) \frac{\eta_{0j}(X)}{p_{0j}}},~ \text{and} \\
r_{1j}(f) & = \prob{f(X)=0 \given Y=1, Z=j} \\ 
& = 1 - \prob{f(X)=1 \given Y=i, Z=j} \\
& = 1 - \expec{X}{f(X) \frac{\eta_{1j}(X)}{p_{1j}}} \\
& = 1 - \expec{X}{f(X)~ u_{1j}(X)}.
\end{align*}
\end{proof}

\subsection{Proof for Proposition \ref{prop:convex-rawls}}
\begin{proof}
For any binary classifier $f: \mathcal{X} \rightarrow \{0, 1\}$,
\begin{align*}
 \max_{i \in \{0, 1\}, j \in [p]}  r_{ij}(f) &= \max_{\substack{\sum_{ij} c_{ij} \leq 1 \\ c_{ij} \geq 0,~ \forall ij}}~  \sum_{i \in \{0, 1\}, j \in [p]} c_{ij} r_{ij}(f) \\
 &= \max_{\substack{\sum_{ij} c_{ij} \leq 1 \\ c_{ij} \geq 0,~ \forall ij}} \expec{X}{f(X) \left(\sum_{j \in [p]} c_{0j} u_{0j}(X) - c_{1j} u_{1j}(X)\right)} + \sum_{j \in [p]} c_{1j},  \hspace{.07\linewidth} \text{using Proposition \ref{prop:error-expec}} \\
 &= \max_{\substack{\sum_{ij} c_{ij} \leq 1 \\ c_{ij} \geq 0,~ \forall ij}}~  \expec{X}{f(X) \left(\sum_{i \in \{0, 1\}, j \in [p]} (-1)^{i} c_{ij} u_{ij}(X)\right)} + \sum_{j \in [p]} c_{1j}.
\end{align*}
\end{proof}

\subsection{Proof for Theorem \ref{thm:rawls}}
\begin{proof}
We use Proposition \ref{prop:convex-rawls} to express the Rawls error rate as minimizing a linear function over binary classifiers $f: \mathcal{X} \rightarrow \{0, 1\}$, where the coefficients of the linear function come from solving an inner maximization. Then we relax our classifier to be real-valued in $[0, 1]$, and use von Neumann's minimax theorem to swap the order of optimization. After the swap, it is easy to see that the optimal relaxed classifier is actually binary-valued, and it must be the Rawls classifier. By Definition \ref{def:rawls} of the Rawls error rate, 
\begin{align*}
& r^{*} = \min_{f: \mathcal{X} \rightarrow \{0, 1\}}~ \max_{i \in \{0, 1\}, j \in [p]} r_{ij}(f) \\
 = & \max_{\substack{\sum_{ij} c_{ij} \leq 1 \\ c_{ij} \geq 0,~ \forall ij}}~ \sum_{i \in \{0, 1\}, j \in [p]} c_{ij} r_{ij}(f) \\
 = & \min_{f: \mathcal{X} \rightarrow \{0, 1\}}~ \max_{\substack{\sum_{ij} c_{ij} \leq 1 \\ c_{ij} \geq 0,~ \forall ij}}~ \expec{X}{f(X) \left(\sum_{ \substack{ i \in \{0, 1\} \\ j \in [p] } } (-1)^{i} c_{ij} u_{ij}(X)\right)} + \sum_{j \in [p]} c_{1j}, \quad \text{using Proposition \ref{prop:convex-rawls}} \\
 \geq & \min_{h: \mathcal{X} \rightarrow [0, 1]}~ \max_{\substack{\sum_{ij} c_{ij} \leq 1 \\ c_{ij} \geq 0,~ \forall ij}}~ \expec{X}{h(X) \left(\sum_{ \substack{ i \in \{0, 1\} \\ j \in [p] } } (-1)^{i} c_{ij} u_{ij}(X)\right)} + \sum_{j \in [p]} c_{1j},~ \text{using relaxation $h: \mathcal{X} \rightarrow [0, 1]$} 
 \end{align*}
 When $u_{ij}(x)$'s are known explicitly through the given distribution $\mathcal{D}$, by von Neumann's minimax theorem in variables $h(x)$ and $c_{ij}$'s, we get
 \begin{align*}
 r* \geq & \max_{\substack{\sum_{ij} c_{ij} \leq 1 \\ c_{ij} \geq 0,~ \forall ij}}~ \min_{h: \mathcal{X} \rightarrow [0, 1]}~ \expec{X}{h(X) \left(  \sum_{ \substack{ i \in \{0, 1\}\\  j \in [p] } } (-1)^{i} c_{ij} u_{ij}(X)\right)} + \sum_{j \in [p]} c_{1j}, \\
= & \max_{\substack{\sum_{ij} c_{ij} \leq 1 \\ c_{ij} \geq 0,~ \forall ij}}~ \expec{X}{h_{c}(X) \left(\sum_{ \substack{ i \in \{0, 1\} \\ j \in [p] }  } (-1)^{i} c_{ij} u_{ij}(X)\right)} + \sum_{j \in [p]} c_{1j}, 
\end{align*}
where $h_{c}(x) = \ind{\sum_{j \in [p]} (-1)^{i} c_{ij} u_{ij}(x) \leq 0}$.
\begin{align*}
r* \geq & \min_{f: \mathcal{X} \rightarrow \{0, 1\}} \max_{\substack{\sum_{ij} c_{ij} \leq 1 \\ c_{ij} \geq 0,~ \forall ij}}~ \expec{X}{f(X) \left(  \sum_{ \substack{ i \in \{0, 1\} \\ j \in [p] } } (-1)^{i} c_{ij} u_{ij}(X)\right)} + \sum_{j \in [p]} c_{1j}, \quad \text{because $h_{c}: \mathcal{X} \rightarrow \{0, 1\}$} \\
 =& \min_{f: \mathcal{X} \rightarrow \{0, 1\}}~ \max_{i \in \{0, 1\}, j \in [p]} r_{ij}(f), \quad \text{using Proposition \ref{prop:convex-rawls}} \\
 =& \; r^{*},
\end{align*}
Therefore, all the above inequalities must be equalities. Moreover, let $c^{*} = \left(c^{*}_{ij}\right)_{i \in \{0, 1\}, j \in [p]}$ be the optimal solution of the above. Hence, the Rawls classifier is given by 
\begin{align*}
f^{*}(x) = h_{c^{*}}(x) & = \ind{\sum_{j \in [p]} (-1)^{i} c^{*}_{ij} u_{ij}(x) \leq 0} = \ind{\sum_{j \in [p]} c^{*}_{1j} u_{1j}(x) - \sum_{j \in [p]} c^{*}_{0j} u_{0j}(x) \geq 0},
\end{align*}
and the Rawls error rate is equal to
\begin{align*}
r^{*} & = \sum_{i \in \{0, 1\}, j \in [p]} c^{*}_{ij} r_{ij}(f^{*}) \\
& = \sum_{i \in \{0, 1\}, j \in [p]} c^{*}_{ij} r_{ij}(h_{c^{*}}) \\
& = \expec{X}{h_{c^{*}}(X) \left(\sum_{i \in \{0, 1\}, j \in [p]} (-1)^{i} c^{*}_{ij} u_{ij}(X)\right)} + \sum_{j \in [p]} c^{*}_{1j} \\
& = \expec{X}{\min\left\{0, \sum_{i \in \{0, 1\}, j \in [p]} (-1)^{i} c^{*}_{ij} u_{ij}(X)\right\}}  + \sum_{j \in [p]} c^{*}_{1j}.
\end{align*}
\end{proof}

\subsection{Proof for Corollary \ref{coro:rawls-non-unique}}
\begin{proof}
Assume that the Rawls classifier $f^{*}$ is non-trivial. By Definition \ref{def:rawls} and the Rawls classifier $f^{*}$ characterized by Theorem \ref{thm:rawls}, the Rawls error rate is equal to 
\[
r^{*} = \max_{i \in \{0, 1\}, j \in [p]} r_{ij}(f^{*}) = \sum_{i \in \{0, 1\}, j \in [p]} c^{*}_{ij} r_{ij}(f^{*}).
\]
Thus, the support of the convex combination given by $c^{*}$ must be only over indices $ij$'s where $r_{ij}(f^{*})$ attains that maximum value. Equivalently, $\{ij \suchthat c^{*}_{ij} > 0\}$ must be the same as $\{ij \suchthat r_{ij}(f^{*}) = r^{*}\}$. Let $ij$ be the unique index such that $c^{*}_{ij} > 0$ and let $c^{*}_{kl} = 0$ for all $(k, l) \neq (i, j)$. Then
\[
f^{*}(x) = \begin{cases} \ind{- c^{*}_{0j} u_{0j}(x) \geq 0} \equiv 0, & \text{if $i=0$}, \\ \ind{c^{*}_{1j} u_{1j}(x) \geq 0} \equiv 1, & \text{if $i=1$}, \end{cases}
\]
because $u_{ij}(x) = \prob{Y=i, Z=j \given X=x}/\prob{Y=i, Z=j} \geq 0$, for all $i \in \{0, 1\}, j \in [p]$ and $x \in \mathcal{X}$. This contradicts the non-triviality of $f^{*}$.
\end{proof}

\subsection{Proof for Corollary \ref{coro:rawls-equal-error-rates}}

\begin{proof}
Assume that the Rawls classifier is non-trivial. Suppose there are no $j, k \in [p]$ such that $S_{0j}$ and $S_{1k}$ attain the Rawls rate $r^{*}$. Then the corresponding $c^{*}$ that characterizes the Rawls classifier $f^{*}$ in Theorem \ref{thm:rawls} must have either $\{ij \suchthat c^{*}_{ij} > 0\} \subseteq \{0j \suchthat j \in [p]\}$ or $\{1j \suchthat j \in [p]\}$. That is, the indices of the non-zero coordinates $c^{*}_{ij}$ must either all have $i=0$ or all have $i=1$. In that case, the corresponding $f^{*}$ by Theorem \ref{thm:rawls} looks like
\[
f^{*}(x) = \begin{cases} \ind{- \sum_{j \in [p]} c^{*}_{0j} u_{0j}(x) \geq 0} \equiv 0, & \text{if $i=0$}, \\ \ind{\sum_{j \in [p]} c^{*}_{1j} u_{1j}(x) \geq 0} \equiv 1, & \text{if $i=1$}, \end{cases},
\]
because $u_{ij}(x) = \prob{Y=i, Z=j \given X=x}/\prob{Y=i, Z=j} \geq 0$, for all $i \in \{0, 1\}, j \in [p]$ and $x \in \mathcal{X}$. This contradicts the non-triviality of $f^{*}$.
\end{proof}

\subsection{Proof for Corollary \ref{coro:rawls-threshold-classifier}}
\begin{proof}
In the $p=1$ case, the Rawls classifier $f^{*}$ in Theorem \ref{thm:rawls} is given by some coefficients $\{c^{*}_{i1}\}_{i \in \{0, 1\}}$, with $\sum_{ij} c^{*}_{ij} = 1$ and $c_{ij} \geq 0$, for all $ij$, such that
\begin{align*}
f^{*}(x) & = \ind{c^{*}_{11} u_{11}(x) - c^{*}_{01} u_{01}(x) \geq 0} \\
& = \ind{\frac{c^{*}_{11}}{p_{11}} \eta_{11}(x) - \frac{c^{*}_{01}}{p_{01}} \eta_{01}(x) \geq 0} \\
& = \ind{\frac{c^{*}_{11}}{p_{11}} \eta_{11}(x) - \frac{c^{*}_{01}}{p_{01}} (1 - \eta_{11}(x)) \geq 0} \\
& = \ind{\left(\frac{c^{*}_{11}}{p_{11}} + \frac{c^{*}_{01}}{p_{01}}\right) \eta_{11}(x) \geq \frac{c^{*}_{01}}{p_{01}}} \\
& = \ind{\eta(x) \geq t},
\end{align*}
where $\eta(x) = \prob{Y=1 \given X=x}$, and the threshold $t$ is equal to
\[
t = \frac{(c_{01}/p_{01})}{(c_{11}/p_{11}) + (c_{01}/p_{01})}.
\]
\end{proof}

\section{Proof details for Section \ref{sec:fair-adapt}}
\label{sec:proof-section-4}

\subsection{Proof for Theorem \ref{thm:fat}}
\begin{proof}
Since we are looking for classifiers $f_{b}(x) = \ind{s(x) \geq b}$, 
\begin{align*}
R_{ij}(f_{b}) & = \sup_{(s, \mathcal{D}) \in \mathcal{R}_{MV}} \prob{f_{b}(s(X) \neq Y \given Y=i, Z=j}  = \begin{cases} \sup_{(s, \mathcal{D}) \in \mathcal{R}_{MV}} \prob{s(X_{0j}) \geq b},~ \text{for $i=0$} \\
\sup_{(s, \mathcal{D}) \in \mathcal{R}_{MV}} \prob{s(X_{1j}) \leq b},~ \text{for $i=1$} \end{cases}
\end{align*}
Therefore, the restricted Rawls error rate is given by the following optimization.
\begin{align*}
& \text{minimize}~ \max_{j \in [p]} R_{j}~ \text{over $b, R_{1}, R_{2}, \dotsc, R_{p}$ subject to} \\
& \sup_{(s, \mathcal{D}) \in \mathcal{R}_{MV}} \prob{s(X_{0j}) \geq b} \leq R_{j},~ \forall j \in [p] \\
& \sup_{(s, \mathcal{D}) \in \mathcal{R}_{MV}} \prob{s(X_{1j}) \leq b} \leq R_{j},~ \forall j \in [p].
\end{align*}
We can equivalently rewrite as
\begin{align*}
& \text{maximize}~ \min_{j \in [p]} (1 - R_{j})~ \text{over $b, R_{1}, R_{2}, \dotsc, R_{p}$ subject to} \\
& \inf_{(s, \mathcal{D}) \in \mathcal{R}_{MV}} \prob{s(X_{0j}) \leq b} \geq 1 - R_{j},~ \forall j \in [p] \\
& \inf_{(s, \mathcal{D}) \in \mathcal{R}_{MV}} \prob{s(X_{1j}) \geq b} \geq 1 - R_{j},~ \forall j \in [p].
\end{align*}
Note that to satisfy $R_{j} \leq 1/2$, for all $j \in [p]$, we must have $\mu_{0j} \leq b \leq \mu_{1j}$, for all $j \in [p]$. Using the upper bound in the one-sided Chebyshev inequality (Proposition \ref{prop:one-sided-chebyshev}) and its tightness property, we can rewrite our optimization problem as follows.
\begin{align*}
& \underset{b, R_{1}, R_{2}, \dotsc, R_{p}}{\text{maximize}}~ \min_{j \in [p]} (1 - R_{j}) \quad \text{subject to} \\
& \frac{(b - \mu_{0j})^{2}}{\sigma_{0j}^{2} + (b - \mu_{0j})^{2}} \geq 1 - R_{j},~ \forall j \in [p] \\
& \frac{(\mu_{1j} - b)^{2}}{\sigma_{1j}^{2} + (\mu_{1j} - b)^{2}} \geq 1 - R_{j},~ \forall j \in [p].
\end{align*}
Equivalently, the above can be written as
\begin{align*}
& \underset{b, R_{1}, R_{2}, \dotsc, R_{p}}{\text{maximize}}~ \min_{j \in [p]} (1 - R_{j}) \quad \text{subject to} \\
& \frac{1}{1 - R_{j}} - 1 \geq \frac{\sigma_{0j}^{2}}{(b - \mu_{0j})^{2}},~ \forall j \in [p] \\
& \frac{1}{1 - R_{j}} - 1 \geq \frac{\sigma_{1j}^{2}}{(\mu_{1j} - b)^{2}},~ \forall j \in [p].
\end{align*}
Further simplification gives
\begin{align*}
& \underset{b, R_{1}, R_{2}, \dotsc, R_{p}}{\text{maximize}}~ \min_{j \in [p]} (1 - R_{j}) \quad \text{subject to} \\
& b \geq \mu_{0j} + \sigma_{0j} \sqrt{\frac{1 - R_{j}}{R_{j}}},~ \forall j \in [p] \\
& \mu_{1j} - \sigma_{1j} \sqrt{\frac{1 - R_{j}}{R_{j}}} \geq b,~ \forall j \in [p].
\end{align*}
Note that the constraints arising from negative square root hold vacuously because $\mu_{0j} \leq b \leq \mu_{1j}$, for all $j \in [p]$. Now we can eliminate $b$ from this optimization as follows.
\begin{align*}
& \underset{R_{1}, R_{2}, \dotsc, R_{p}}{\text{maximize}}~ \min_{j \in [p]} (1 - R_{j}) \quad \text{subject to} \\
& \mu_{1j} - \sigma_{1j} \sqrt{\frac{1 - R_{j}}{R_{j}}} \geq \mu_{0j} + \sigma_{0j} \sqrt{\frac{1 - R_{j}}{R_{j}}},~ \forall j \in [p].
\end{align*}
Equivalently, this can be written as
\begin{align*}
& \underset{R_{1}, R_{2}, \dotsc, R_{p}}{\text{maximize}}~ \min_{j \in [p]} (1 - R_{j}) \quad \text{subject to} \\
& \frac{\mu_{1j} - \mu_{0j}}{\sigma_{1j} + \sigma_{0j}} \geq \sqrt{\frac{1 - R_{j}}{R_{j}}},~ \forall j \in [p].
\end{align*}
Since $\sqrt{(1 - a)/a}$ is a monotonically decreasing function for $a > 0$, we need to only find
\[
j^{*} = \underset{j \in [p]}{\text{argmin}}~ \frac{\mu_{1j} - \mu_{0j}}{\sigma_{1j} + \sigma_{0j}}.
\]
Thus, the optimal $\eta_{j^{*}}$ is
\[
\eta_{j^{*}} = \left(1 + \left(\frac{\sigma_{1j^{*}} + \sigma_{0j^{*}}}{\mu_{1j^{*}} - \mu_{0j^{*}}}\right)^{2}\right)^{-1},
\]
and the optimal threshold $b^{*}$ is
\begin{align*}
b^{*} & = \mu_{1j^{*}} - \sigma_{1j^{*}} \sqrt{\frac{\mu_{1j^{*}} - \mu_{0j^{*}}}{\sigma_{1j^{*}} + \sigma_{0j^{*}}}}  = \mu_{0j^{*}} + \sigma_{0j^{*}} \sqrt{\frac{\mu_{1j^{*}} - \mu_{0j^{*}}}{\sigma_{1j^{*}} + \sigma_{0j^{*}}}}.
\end{align*}
\end{proof}

Here we state the one-sided Chebyshev inequality for completeness. Its bound on the deviation away from the mean of a random variable plays a key role in our proof.
\begin{prop} \label{prop:one-sided-chebyshev} 
For any real-valued random variable $X$ with mean $\mu$ and standard deviation $\sigma$, and for any $a > 0$, 
\begin{align*}
\prob{X \geq \mu + a} & \leq \frac{\sigma^{2}}{\sigma^{2} + a^{2}} \\
\prob{X \leq \mu - a} & \leq \frac{\sigma^{2}}{\sigma^{2} + a^{2}}.
\end{align*}
Moreover, for any given $a > 0$, there exists a distribution $X$ with mean $\mu$ and standard deviation $\sigma$ such that these inequalities are tight. 
\end{prop}

\subsection{Proof for Theorem \ref{thm:flat}}
\begin{proof}
Let $s(X_{ij})$ be a Gaussian with mean $\mu_{ij} \in \R^{d}$ and covariance matrix $\Sigma_{ij} \in \R^{d \times d}$, respectively, for $i \in \{0, 1\}$ and $j \in [p]$. Let $f_{w,b}$ denote a linear classifier given by $f_{w,b}(x) = \ind{w^{T} s(x) \geq b}$, for some $w \in \R^{d}$ and $b \in \R$. Then, for $i \in \{0, 1\}$ and $j \in [p]$, we get
\begin{align*}
\prob{f_{w,b}(X_{ij})=0} & = \prob{w^{T} s(X_{ij}) \leq b} = \Phi\left(\frac{b - w^{T}\mu_{ij}}{\norm{\Sigma_{ij}^{1/2} w}}\right),
\end{align*}{}

where $\Phi(\cdot)$ is the CDF of a standard normal variable $N(0, 1)$. Similarly, using $\Phi(t) + \Phi(-t) = 1$, we have

\begin{align*}
\prob{f_{w,b}(X_{ij})=1} &= \prob{w^{T} s(X_{ij}) \geq b} = \Phi\left(\frac{w^{T}\mu_{ij} - b}{\norm{\Sigma_{ij}^{1/2} w}}\right).
\end{align*}{}

Thus, assuming that $s(X_{ij})$'s are Gaussians with the given means and covariance matrices, the restricted Rawls error rate can be obtained by solving the following optimization problem (where error rate constraints are replaced by accuracy constraints).
\begin{align*}
& \underset{w, b, r_{1}, r_{2}, \dotsc, r_{p}}{\text{maximize}}~ \min_{j} (1 - r_{j}) \quad \text{subject to} \\
& \Phi\left(\frac{b - w^{T}\mu_{0j}}{\norm{\Sigma_{0j}^{1/2} w}}\right) \geq 1 - r_{j} \quad \text{and} \hspace{0.5cm} \Phi\left(\frac{w^{T}\mu_{1j} - b}{\norm{\Sigma_{1j}^{1/2} w}}\right) \geq 1 - r_{j},~ \forall j \in [p].
\end{align*}
\begin{align*}
& \text{In other words}, \quad \underset{w, b, r_{1}, r_{2}, \dotsc, r_{p}}{\text{maximize}}~ \min_{j} (1 - r_{j}) \quad \text{subject to} \\
& b \geq w^{T}\mu_{0j} + \Phi^{-1}(1 - r_{j})~ \norm{\Sigma_{0j}^{1/2} w} \quad \text{and} \hspace{0.5cm} w^{T}\mu_{1j} - \Phi^{-1}(1 - r_{j})~ \norm{\Sigma_{1j}^{1/2} w} \geq b,~ \forall j \in [p].
\end{align*}
Observe that $b$ can be eliminated in the above to get
\begin{align*}
& \underset{w, r_{1}, r_{2}, \dotsc, r_{p}}{\text{maximize}}~ \min_{j} (1 - r_{j}) \quad \text{subject to} \\
& w^{T}\mu_{0j} + \Phi^{-1}(1 - r_{j})~ \norm{\Sigma_{0j}^{1/2} w} \leq w^{T}\mu_{1j} - \Phi^{-1}(1 - r_{j})~ \norm{\Sigma_{1j}^{1/2} w}, \forall j \in [p].
\end{align*}
Moreover, at the optimum $(w_{*}, b_{*}, r_{1}^{*}, r_{2}^{*}, \dotsc, r_{p}^{*})$, the constraint must be tight for some $j^{*} \in [p]$, giving

\begin{align*}
b_{*} & = w_{*}^{T} \mu_{0j^{*}} + \Phi^{-1}(1 - r_{{j}^{*}})~ \norm{\Sigma_{0j^{*}}^{1/2} w_{*}} \\
& = w_{*}^{T} \mu_{1j^{*}} - \Phi^{-1}(1 - r_{j}^{*})~ \norm{\Sigma_{1j^{*}}^{1/2} w_{*}}.
\end{align*}{}

Using change of variables $\kappa_{j} = \Phi^{-1}(1 - r_{j})$ and monotonicity of $\Phi(\cdot)$, we can rewrite our optimization as follows.
\begin{align*}
& \underset{w, \kappa_{1}, \kappa_{2}, \dotsc, \kappa_{p}}{\text{maximize}}~ \min_{j} \kappa_{j} \quad \text{subject to} \\
& w^{T}(\mu_{1j} - \mu_{0j}) \geq \kappa_{j} \left(\norm{\Sigma_{1j}^{1/2} w} + \norm{\Sigma_{0j}^{1/2} w}\right),~ \forall j \in [p].
\end{align*}
We can eliminate $\kappa_{j}$'s from this to write it as an equivalent optimization problem only over $w$.
\begin{align*}
& \underset{w}{\text{maximize}}~ \min_{j} \frac{w^{T}(\mu_{1j} - \mu_{0j})}{\norm{\Sigma_{1j}^{1/2} w} + \norm{\Sigma_{0j}^{1/2} w}},~ \text{equivalently}, \\
& \underset{w}{\text{minimize}}~ \max_{j} \frac{\norm{\Sigma_{1j}^{1/2} w} + \norm{\Sigma_{0j}^{1/2} w}}{w^{T}(\mu_{1j} - \mu_{0j})}.
\end{align*}
If $\mu_{1j} = \mu_{0j}$, for any $j \in [p]$, then the optimum $\kappa^{*}$ must be zero. Otherwise, observe that the above constraints are homogeneous in $w$, i.e., whenever $w$ satisfies these constraints, any non-negative scalar multiple of $w$ also satisfies them. It is known that the optimization problem mentioned above can be solved efficiently in polynomial time using SOCP (Second Order Cone Programming) \cite{Alizadeh2003} and fractional programming \cite{Schaible1983}. A general framework to solve such problem is discussed in \cite{LanckrietGBJ2002}, which we defer to the full version of this paper. For now, it can be seen that when the covariance matrices are diagonal, this problem can be solved using SOCP.

We can assume $p \leq d$, that is the number of sensitive groups (e.g., race, gender) is smaller than the dimensionality of the feature space. In this case, WLOG, we can rewrite the above optimization as
\begin{align*}
& \underset{w}{\text{minimize}}~ \max_{j} \norm{\Sigma_{1j}^{1/2} w} + \norm{\Sigma_{0j}^{1/2} w} \quad \text{subject to} \\
& w^{T} (\mu_{1j} - \mu_{0j}) = 1,~ \forall j \in [p].
\end{align*}

There is a nice geometric interpretation for this optimal solution in the case of spherical Gaussians, i.e., when each covariance matrix $\Sigma_{ij} = \sigma_{ij}^{2} I$, for some $\sigma_{ij} \geq 0$ and the identity matrix $I \in \R^{d \times d}$. Then the above optimization becomes
\[
\underset{w}{\text{minimize}}~ \max_{j} \frac{\norm{w} (\sigma_{1j} + \sigma_{0j})}{w^{T}(\mu_{1j} - \mu_{0j})}.
\]
Or equivalently,
\begin{align*}
& \underset{w}{\text{minimize}}~ \norm{w}~ \text{subject to} \\
& w^{T}(\mu_{1j} - \mu_{0j}) \geq \sigma_{1j} + \sigma_{0j},~ \forall j \in [p].
\end{align*}
Geometrically, this corresponds to the following classical problem \cite{Wolfe1976}: given a polyhedron, find a point on it that is closest to the origin in $\ell_{2}$ norm. This is a special case of convex quadratic optimization and is known to be solvable efficiently in polynomial time using the Ellipsoid method as well as the interior point method \cite{Boyd2004}, and its complexity is similar to that of the support vector machine problem.
\end{proof}

\end{document}